\documentclass[conference]{IEEEtran}
\IEEEoverridecommandlockouts
\usepackage{cite}
\usepackage{amsmath,amssymb,amsfonts}
\usepackage{graphicx}
\usepackage{textcomp}
\usepackage[hidelinks]{hyperref}
\usepackage{booktabs,multicol, multirow}
\usepackage{xcolor}
\usepackage{adjustbox}
\usepackage[utf8]{inputenc}
\usepackage[T1]{fontenc}
\usepackage{algorithm}
\usepackage{algpseudocode}

\def\BibTeX{{\rm B\kern-.05em{\sc i\kern-.025em b}\kern-.08em
    T\kern-.1667em\lower.7ex\hbox{E}\kern-.125emX}}
\begin{document}

\title{Deep Emotions Across Languages: \\A Novel Approach for Sentiment Propagation \\in Multilingual WordNets}

\author{\IEEEauthorblockN{Jan Kocoń}
\IEEEauthorblockA{\textit{Department of Artificial Intelligence}\\ \textit{Wrocław University of Science and Technology, Poland} \\
jan.kocon@pwr.edu.pl}
}

\maketitle

\begin{abstract}
 Sentiment analysis involves using WordNets enriched with emotional metadata, which are valuable resources. However, manual annotation is time-consuming and expensive, resulting in only a few WordNet Lexical Units being annotated. This paper introduces two new techniques for automatically propagating sentiment annotations from a partially annotated WordNet to its entirety and to a WordNet in a different language: Multilingual Structured Synset Embeddings (MSSE) and Cross-Lingual Deep Neural Sentiment Propagation (CLDNS). We evaluated the proposed MSSE+CLDNS method extensively using Princeton WordNet and Polish WordNet, which have many inter-lingual relations. Our results show that the MSSE+CLDNS method outperforms existing propagation methods, indicating its effectiveness in enriching WordNets with emotional metadata across multiple languages. This work provides a solid foundation for large-scale, multilingual sentiment analysis and is valuable for academic research and practical applications.
\end{abstract}


\section{Introduction}

 Sentiment analysis, the process of analyzing human emotions through text, has become increasingly popular in the age of widespread online communication. Whether it's consumer reviews, social media conversations, or news articles, recognizing and interpreting emotional tone automatically is no longer just a technological feat but a necessity for society \cite{ramteke2016,cambria2017affective,kocon2019multilingual,Kocon2019multi,kocon2019multi2,kocon2019recognition,kanclerz2020cross,wierzba2021emotion,kocon2021aspectemo,kocon2021multiemo,cambria2022senticnet,baran2022linguistic}. Although deep learning techniques such as BERT \cite{devlin2018bert,korczynski2022compression,kocon2022neuro,srivastava2023beyond} have significantly improved the field, achieving state-of-the-art results in various NLP tasks, the need for high-quality, annotated resources remains a crucial factor for performance \cite{dashtipour2016multilingual,kocon2021mapping,milkowski2021personal,milkowski2022multitask}. This is especially challenging for under-resourced languages and cross-lingual sentiment analysis tasks.

 Big Data has raised concerns about methodology and scalability. While large language models like GPT-3 and GPT-4 are impressive, they are not practical for specialized tasks like sentiment analysis. These models require significant resources and cannot replace high-quality, domain-specific training data \cite{kocon2023chatgpt}. Additionally, their monolithic size and one-size-fits-all approach are not efficient for customized solutions like sentiment propagation in multilingual WordNets.

\begin{figure}
\centerline{\includegraphics[width=\linewidth]{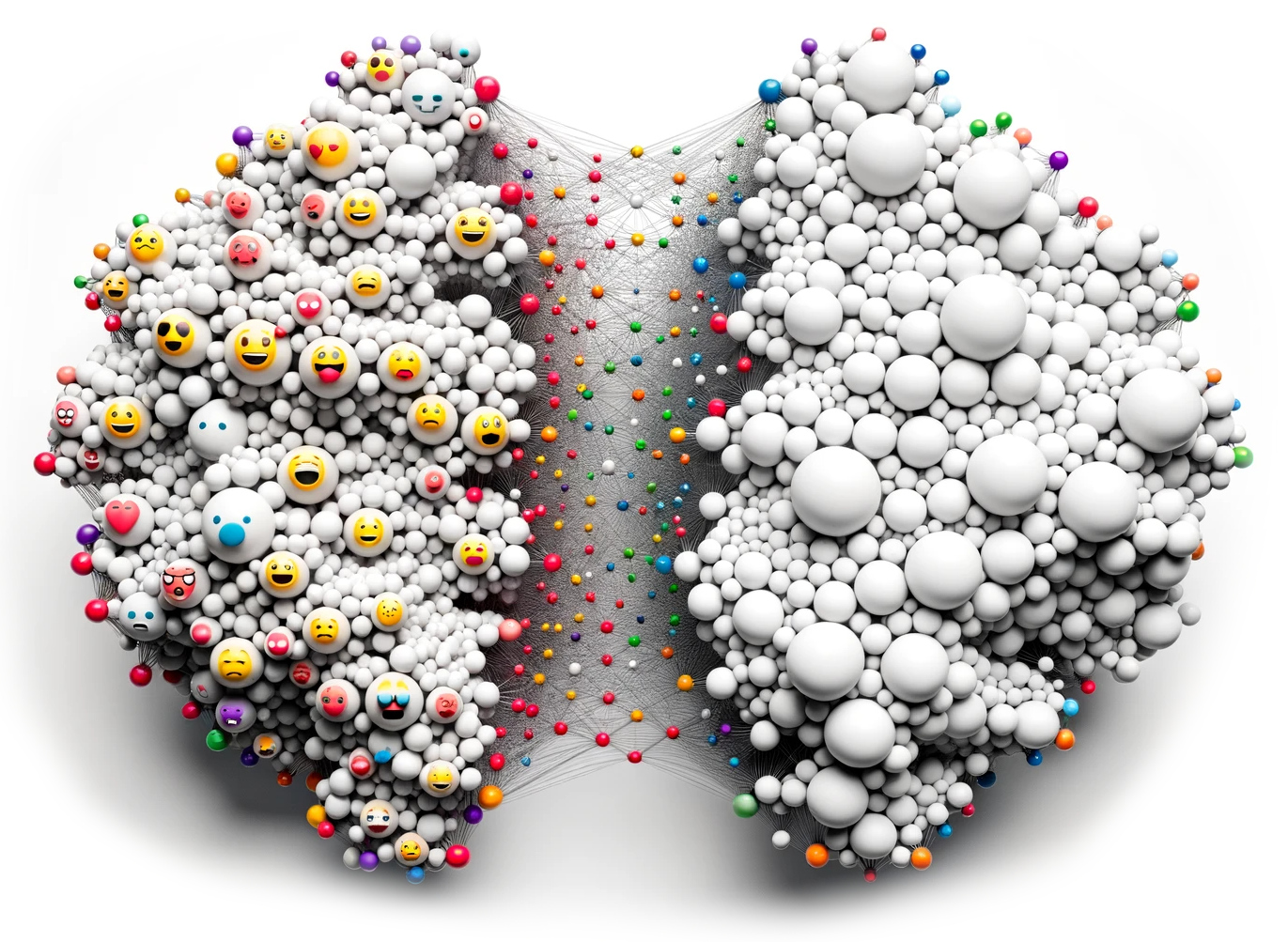}}
\caption{Propagation of the emotional annotation from one WordNet to another. Example WordNets are connected at the level of synsets and lexical units with inter-lingual relations (Fig. generated with the assistance of DALL·E 3). }
\label{fig:emo_prop}
\end{figure}

 In this paper, we introduce two new techniques, Multilingual Structured Synset Embeddings (MSSE) and Cross-Lingual Deep Neural Sentiment Propagation (CLDNS), which aim to address existing gaps in sentiment annotations. These techniques help propagate sentiment annotations from a partially annotated WordNet to its entirely unannotated counterpart and a WordNet in a different language. We thoroughly evaluated the techniques using Princeton WordNet and Polish WordNet, which share a significant set of inter-lingual relations. The results showed that MSSE+CLDNS outperforms current sentiment propagation methods significantly. This improvement leads to enriching WordNets across multiple languages with high-quality emotional metadata.

Our study offers an overview of sentiment analysis resources for the Polish language. We assessed current WordNet sentiment methods and contrasted them with our new approaches. Our main contributions are: (1) introducing a new embedding technique, MSSE, capturing human emotions in graph-based databases; (2) presenting CLDNS, a method for emotion propagation in multilingual WordNets using MSSE; (3) aiming to harmonize modern NLP computational needs with the detailed demands of sentiment analysis in different languages.

\section{Related Work}
\label{sec:related-work}

Our work builds upon previous research in sentiment analysis, cross-lingual sentiment propagation, and knowledge base embeddings \cite{HabernalBrychcin2013b,he2016ups,dashtipour2016multilingual,Bucar2018,rogers2018rusentiment}. Traditionally, there have been three approaches for creating sentiment lexicons: corpus-based, dictionary-based, and manual annotation \cite{Liu2015}. However, manual annotation tends to be expensive, which has led to the development of automatic techniques to expand lexicons.

 Recent work has specifically targeted the propagation of sentiment information in WordNet-like structures \cite{Maks2011,Esuli:2006,janz2017,Kocon2018,Kocon2019a}. It's worth noting that current efforts in the field of sentiment analysis have primarily focused on Lexical Units (LUs) or synsets, and have been limited to a single language. In contrast, our research aims to achieve cross-lingual sentiment propagation through the use of Multilingual Structured Synset Embeddings (MSSE) and Cross-Lingual Deep Neural Sentiment Propagation (CLDNS). These methods are designed to work across multiple languages and more accurately capture the broader semantic context of the text.

Embedding techniques are crucial for NLP tasks like sentiment analysis \cite{bojanowski2017enriching,devlin2018bert,milkowski2022multi}. The latest advances in NLP, such as Transformer models and large pre-trained models like BERT \cite{devlin2018bert}, GPT-2, GPT-3, ChatGPT, and GPT-4 \cite{radford2019language, brown2020language,kazienko2023human,kocon2023chatgpt,koptyra2023clarin,kocon2023differential,ferdinan2023personalized,mieleszczenko2023capturing}, have set new benchmarks for a variety of tasks. 
 Although large language models have proven to be highly efficient, their size and computational demands make them less appropriate for certain tasks, such as propagating sentiment across WordNets. Therefore, our approach seeks to overcome the limitations of traditional WordNet-based methods and large pre-trained models by combining the benefits of structured embeddings and deep learning techniques.

 With the advent of large language models, there is now an opportunity to reconsider the way we utilize and expand knowledge bases. These models can be used to generate fresh knowledge and supplement existing resources, which makes it possible to automatically enhance WordNet-like lexicons \cite{petroni2019language, bosselut2020dynabench}. Typically, the techniques employed to develop models tend to be very resource-intensive, which makes them challenging to fine-tune for specific tasks like spreading sentiment across different languages \cite{kocon2023chatgpt}. Our MSSE+CLDNS approach aims to bridge this gap in the literature by combining the power of deep models with the efficiency and specificity of structured embeddings.

 An increasing trend in sentiment analysis is to move beyond simplistic sentiment polarity and instead analyze the underlying emotions. \cite{mohammad2018semeval}. The work on EmoReact \cite{felbo2017using} and GoEmotions \cite{demszky2020goemotions} extends this further by mapping text to multiple dimensions of emotions, much like our work, which considers not just polarity but also dimensions of arousal and Plutchik's eight basic emotions \cite{Plutchik}.

 The continuous need for computational power of deep learning models is a major concern, as discussed in \cite{strubell2019energy}. This issue becomes especially challenging in a cross-lingual setting, as each additional language requires a model of similar size to be trained, making the scaling problem more severe. Our approach aims to address this by utilizing efficient graph-based techniques to propagate sentiment information across different languages, without the need for complex models.

 Our work combines different research areas to advance sentiment analysis across multiple languages. We use structured embeddings and neural network architectures designed for this task. Our approach introduces innovative methods for cross-lingual sentiment propagation, providing a more efficient alternative to large language models. 

\section{Multilingual Structured Synset Embeddings}
\label{sec:msse}

 This paper proposes a solution to the problem of limited emotional data in WordNet LUs by introducing Multilingual Structured Synset Embeddings (MSSE). MSSE is an advanced embedding technique that not only captures lexical semantics but also incorporates structural information of the WordNet graph (see Algorithm~\ref{alg:MSSE}). Our methodology is an extension and refinement of prior methods \cite{goikoetxea2015random, kocon2019recognition}, enriched to handle multilingual WordNets, specifically Princeton WordNet \cite{Miller1995} and Polish WordNet \cite{piasecki2009wordnet}.

\begin{algorithm}[ht]
\caption{MSSE}
\label{alg:MSSE}
\begin{algorithmic}[1]
\Require{WordNet graph $G$, Number of dimensions $d$, Random walk length $l$, Number of random walks $n$, Skip-gram window size $w$}
\Ensure{Embeddings for WordNet nodes and edges}

\Function{GenerateCorpus}{$G, I$}
  \State Initialize an empty corpus $C$
  \For{$i$ in range($n$)}   
    \State Randomly select a starting node $n$ in $G$
    \State Perform a random walk of fixed length $l$, recording a sequence of all visited \emph{node IDs} and \emph{edge IDs} , ensuring each node in a sequence is visited only once
    \State Add the sequence to $C$ 
  \EndFor
  \State \Return{$C$}
\EndFunction

\Function{TrainEmbeddings}{$C, d$}
  \State Initialize empty embedding model
  \For{each node sequence in $C$}
    \State Update the embedding model using Skip-gram with context size $w$
  \EndFor
  \State \Return{Embeddings for WordNet nodes and edges}
\EndFunction
\end{algorithmic}
\end{algorithm}

 MSSE stands out from other approaches for its unique training methodology. Unlike other models, it doesn't rely on pre-existing models like cross-lingual FastText embeddings \cite{grave2018learning}. Instead, it is trained from scratch, using only the textual representations of the WordNet nodes and edges. To achieve multilingual alignment, we leverage the dense inter-lingual relations between Princeton WordNet and Polish WordNet \cite{rudnicka2017mapping,rudnicka2019sense,rudnicka2021non}. To conduct random walks across both WordNets, we utilize a Skip-gram algorithm \cite{guthrie2006closer}. This neural architecture typically generates word embeddings by predicting local context based on a large textual corpus. The model assumes that semantically related words often occur in similar contexts, allowing the embeddings to capture such relationships \cite{Mikolov2013}. However, in contrast to the standard Skip-gram approach, we create an artificial corpus from WordNet, based on Goikoetxea et al. work \cite{goikoetxea2015random}. This corpus is formed through random walks on the WordNet graph, which captures the subtle interplay between synsets (groups of cognitive synonyms), lexical units, and relation types. The embedding of each node is determined by the context along the random walk path. Nodes that are inter-lingual synonyms of each other often appear in close proximity within the context window, which leads to the creation of similar embeddings. This illustrates the potential of MSSE to uncover meaningful multilingual parallels.

 A major innovation in MSSE is incorporating inter-WordNet relations, a feature largely overlooked in previous studies \cite{goikoetxea2015random}. The dense links between Princeton WordNet and Polish WordNet, which were developed under the CLARIN-PL project \cite{piasecki2009wordnet}, are crucial. Their extensive inter-connectedness allows us to perform random walks within a single WordNet and across different language-specific WordNets. Furthermore, MSSE includes the type of relation that connects different units or synsets, adding another layer of semantic information to the embeddings.

 Our artificial corpus includes identifiers for synsets, lexical units, and various relation types. For example, relation types are labeled with prefixes such as "rSS" for synset-to-synset relations or "rSL" for synset-to-lexical unit relations, similarly as in \cite{Kocon2019a}. Additional markers indicate the specific relation type, such as hyponymy or hypernymy. This approach allows for a more detailed analysis of embeddings, which become semantically rich, capturing the concepts and their intricate relationships within and across languages.

 We use the FastText algorithm \cite{bojanowski2017enriching} to create embeddings. This algorithm is known for its effectiveness in various NLP tasks, including sentiment analysis. Our setup generates 300-dimensional embeddings that capture various lexical and emotional nuances. This high-dimensional representation is more informative and versatile, making it suitable for complex tasks like sentiment propagation across multilingual WordNets.

We chose FastText over other options like Word2Vec\cite{church2017word2vec} and GloVe \cite{pennington2014glove} due to its unique ability to capture subword information. FastText represents words using character n-grams, allowing it to handle out-of-vocabulary words and effectively capture variations in word structure. This feature is especially useful in multilingual contexts where languages have diverse morphological structures. Furthermore, FastText has demonstrated strong performance in various NLP tasks \cite{grave2018learning,kocon2019evaluating,laippala2019toward,aguero2021deep}, making it a reliable choice for our multilingual sentiment propagation framework.

 MSSE is a reliable and efficient multilingual tool for large-scale sentiment analysis. It captures words' emotional undertones and lexical semantics, making it an ideal option for sentiment analysis. When combined with our CLDNS, MSSE far outperforms other sentiment propagation methods.

\section{Cross-Lingual Deep Neural Sentiment Propagation}
\label{sec:cldnsp}
In this section, we will discuss Cross-Lingual Deep Neural Sentiment Propagation (CLDNSP), a mechanism designed to automatically propagate sentiment annotations across different WordNets using the framework of Multilingual Structured Synset Embeddings (MSSE). We will explore the intricacies of CLDNSP, a  method for automating the propagation of sentiment annotations across WordNets of different languages.

We address the limitations of previous methods that utilized a simple logistic regression \cite{Kocon2019a} by introducing a deep neural network-based multilabel classifier. Specifically, we propose two configurations:
\begin{enumerate}
    \item \emph{Base Configuration} is based on \cite{Kocon2019a} and consists of a two-layer neural network: an input layer with 300 neurons and an output layer comprising 26 neurons to capture the multiple dimensions of sentiment.
    \item \emph{Deep Configuration} builds on the Base by adding multiple layers and dropout for regularization, configured as follows: an initial dense layer with 300 inputs and 4096 outputs, followed by a 20\% dropout layer; a subsequent dense layer with 1024 neurons and another 20\% dropout layer; a penultimate dense layer with 256 neurons; and finally, an output layer with 26 neurons.
\end{enumerate}
In both configurations, we employ the ReLU activation function for hidden layers and a linear activation function for the output layer. We use the coefficient of determination as a loss function and early stopping with patience equal to 30. 

 The proposed methodology advances multilingual sentiment analysis research by efficiently transferring sentiment annotation between WordNets. It eliminates the need for manual annotations and provides a scalable approach to enrich WordNets with emotional metadata across languages. The CLDNSP method leverages the inter-lingual richness offered by MSSE and adopts a deep neural architecture for sentiment propagation, taking us closer to achieving automated, fine-grained emotional understanding across languages. Our preliminary results suggest that we are on a promising trajectory toward mapping the complex landscape of human emotions across languages.

\section{Emotional Annotation in Polish Wordnet}
\label{sec:dataset}

For the experiments, we decided to use the plWordNet 3.0 dataset \cite{zasko2018towards}. The emotional annotation process was conducted to enhance the lexical resource with emotive lexicons, which included sentiment polarities, fundamental human values, and basic emotions. This was done to improve its application in emotional and sentiment analysis. The annotations were manually done by linguists and psychologists while following consistent guidelines. These guidelines included the usage of corpora, substitution tests, and lexical unit properties, as presented in \cite{zasko2018towards}. The annotation dimensions that were used included:
\begin{itemize}
    \item \textbf{Sentiment Polarity (pol):} Encoded on a 6-grade scale distinguishing between strong and weak polarisation: Strong Positive, Weak Positive, Strong Negative, Weak Negative, Ambivalent, Neutral
    \item \textbf{Basic Emotions (emo):} Based on Plutchik's framework \cite{Plutchik}, including Joy, Fear, Surprise, Sadness, Disgust, Anger, Trust, and Anticipation.
    \item \textbf{Fundamental Human Values (val):} Influenced by Puzynina's work \cite{puzynina1992jkezyk} and consists of: Non-usefulness, Mistake, Ugliness, Goodness, Harm, Ignorance, Unhappiness, Beauty, Truth, Happiness, Usefulness, and Knowledge.
\end{itemize}

Over 31,000 LUs were annotated in plWordNet 4.0 emo. The inter-annotator agreement for sentiment polarity was notably high, encompassing negative (strong and weak), neutral, and positive (strong and weak) polarities. Negative emotions and values were predominant, aligning with trends in other sentiment lexicons. Notably, there are almost 300k manually annotated interlingual relations between Princeton WordNet and plWordNet \cite{rudnicka2021non}.

\section{Evaluation}
\label{sec:evaluation}

Our evaluation framework is unique in its approach, focusing on multiple levels of WordNet concepts and employing a robust and comprehensive MSSE embedding method to obtain common embeddings for densely connected plWordNet and Princeton WordNet. As a reference embedding method, we used Heterogeneous Structured Synset Embeddings (HSSE) described in \cite{Kocon2019a} with embeddings trained using plWordNet only (with no connections to Princeton WordNet). The propagation procedure is the same as in \cite{Kocon2019a}, but instead of logistic regression, we used CLDNSP in two variants (Base and Deep, see Section~\ref{sec:cldnsp}). 

We utilize 10-fold cross-validation and propagate emotional metadata from an initial seed (train set) that contains 80\% of annotated lexical units (LUs) in plWordNet; a validation set containing 10\% of LUs (for early stopping) and 10\% in the test set. Our propagation method adapts the approach from \cite{Kocon2018}. At first, we train the CLDNSP classifier on an initial seed of LUs, where the input for the model is the LU embedding (obtained via HSSE or MSSE), and the output is the multidimensional emotional annotation of the LU. Then, we apply the classifier to all nearest neighbors of seed LUs and repeat the procedure until we reach all LUs in the test set. Finally, we calculate the precision, recall, and F1 score on a test set and repeat the whole procedure on another set of folds. In the end, we have a sample of 10 results. The code is available in GitHub repository\footnote{\url{https://github.com/KoconJan/deeprop}}, and the data can be obtained here\footnote{\url{http://plwordnet.pwr.edu.pl/wordnet}} upon request.

\subsection{Statistical Significance Tests}

 We use the Shapiro-Wilk test to determine if our results are normally distributed. Through this test, we observe p-values higher than the level of significance ($\alpha=0.05$) for each set of folds in our 10-fold cross-validation. As a result, we fail to reject the null hypothesis that our results are normally distributed. We then use the paired-differences Student's t-test to check the statistical significance of the observed differences, using a significance level of $\alpha=0.05$ \cite{dietterich1998approximate}.

\section{Results}
\label{sec:results}

\subsection{Comparison of Configurations}

 Table~\ref{tab:results1} presents a comparative analysis of four different configurations employing combinations of embeddings (MSSE or HSSE) and classifiers (Base or Deep). These configurations are denoted as HB, HD, MB, and MD:

\begin{itemize}
    \item HB -- HSSE with Base classifier \cite{Kocon2019a},
    \item HD -- HSSE with Deep classifier (CLDNSP),
    \item MB -- MSSE with Base classifier \cite{Kocon2019a},
    \item MD -- MSSE with Deep classifier (CLDNSP).
\end{itemize}

Our statistical tests show that there is no significant difference between the HD and MB configurations. However, both of these configurations are significantly different from the HB configuration. This implies that either the deep classifier or the MSSE method (or both) contribute to enhanced performance. The MD configuration yields better F1-score values, for almost all emotional dimensions.

\begin{table}[h]
 \begin{center}
 \caption{Comparison of the results between different configurations (H - HSSE, M - MSSE, B - Base, D - Deep (CLDNSP)). Measures used are R and R-squared ($R^2 = 1 - \textit{FVU}$).  \label{tab:results1}}
 \begin{adjustbox}{width=\columnwidth,center}
\begin{tabular}{|ccccccccc|}
      \hline
     Measure & & HB & & HD & & MB & & MD \\ \hline
    $R$ & & 0.628$\pm$0.008 & & 0.641$\pm$0.015 & & 0.651$\pm$0.007 & & \textbf{0.693$\pm$0.007} \\
    $R^2$ & & 0.394$\pm$0.010 & & 0.412$\pm$0.019 & & 0.423$\pm$0.009 & & \textbf{0.480$\pm$0.010} \\     
      \hline
\end{tabular}
\end{adjustbox}

 \end{center}
\end{table}

\subsection{Impact of Cross-Lingual Information}

 In Table~\ref{tab:results2}, we compare the HD and MD configurations to assess the impact of cross-lingual information. Emotional dimensions are described in Section~\ref{sec:dataset}. The results show a consistent improvement in performance across all sentiment, emotional, and valuation categories for the MD setup. This highlights the significance of interlingual relations between the Polish and English synsets, emphasizing the importance of cross-lingual information.

\begin{table}
\centering
\caption{
     Comparison of the results between HD  and MD. \textbf{Bold} values indicate the best performance among all architectures.\label{tab:results2}
}
\begin{adjustbox}{width=\columnwidth}
\begin{tabular}{lcccccc}
    \toprule
                        & \multicolumn{3}{c}{HD} & \multicolumn{3}{c}{MD}  \\
    \multicolumn{1}{c}{Emotion}    & P           & R           & F1             & P                & R             & F1 \\
    \midrule
    val\_nonusefulness    & 67.6$\pm$0.9  & 70.0$\pm$1.2  &  68.8$\pm$0.8  &  68.0$\pm$1.0  &  72.2$\pm$1.9  &  \textbf{69.8$\pm$0.4}    \\     
    val\_mistake          & 62.8$\pm$2.6  & 64.8$\pm$1.5  &  63.6$\pm$1.1  &  64.2$\pm$1.6  &  64.8$\pm$3.3  &  \textbf{64.4$\pm$1.1}    \\     
    val\_ugliness         & 65.4$\pm$3.7  & 47.6$\pm$3.2  &  \textbf{55.0$\pm$3.5}  &  67.2$\pm$3.4  &  45.6$\pm$3.0  &  54.4$\pm$3.0    \\     
    emo\_anticipation     & 42.0$\pm$2.7  & 31.0$\pm$3.2  &  35.2$\pm$1.8  &  37.2$\pm$2.3  &  35.2$\pm$3.1  &  \textbf{36.0$\pm$1.2}    \\     
    val\_goodness         & 47.0$\pm$2.3  & 46.0$\pm$2.2  &  46.2$\pm$1.3  &  48.2$\pm$6.9  &  48.2$\pm$3.3  &  \textbf{48.0$\pm$3.7}    \\     
    val\_harm             & 59.8$\pm$2.0  & 64.6$\pm$1.1  &  62.0$\pm$1.4  &  61.0$\pm$3.1  &  66.2$\pm$3.0  &  \textbf{63.4$\pm$1.5}    \\     
    val\_ignorance        & 59.2$\pm$5.8  & 37.0$\pm$4.8  &  45.4$\pm$5.0  &  60.4$\pm$3.6  &  36.6$\pm$6.2  &  45.4$\pm$5.0    \\     
    val\_unhappiness      & 55.4$\pm$1.3  & 60.2$\pm$1.9  &  57.6$\pm$0.9  &  55.6$\pm$2.6  &  62.2$\pm$3.1  &  \textbf{58.8$\pm$1.8}    \\     
    val\_beauty           & 63.4$\pm$2.3  & 46.2$\pm$3.5  &  53.4$\pm$2.1  &  63.4$\pm$5.8  &  48.0$\pm$4.7  &  \textbf{54.4$\pm$3.6}    \\     
    val\_truth            & 51.0$\pm$2.8  & 28.6$\pm$3.1  &  36.6$\pm$2.9  &  51.6$\pm$6.7  &  33.8$\pm$4.9  &  \textbf{40.4$\pm$3.5}    \\     
    emo\_joy              & 67.4$\pm$0.5  & 60.6$\pm$1.7  &  64.0$\pm$0.7  &  67.2$\pm$1.8  &  63.8$\pm$1.8  &  \textbf{65.4$\pm$1.1}    \\     
    emo\_sadness          & 62.4$\pm$1.5  & 69.0$\pm$0.7  &  65.2$\pm$0.8  &  63.4$\pm$1.1  &  70.2$\pm$1.5  &  \textbf{66.6$\pm$0.9}    \\     
    val\_happiness        & 58.0$\pm$1.7  & 58.4$\pm$2.3  &  58.0$\pm$1.4  &  57.6$\pm$3.0  &  60.6$\pm$2.7  &  \textbf{59.0$\pm$1.6}   \\     
    emo\_fear             & 52.8$\pm$3.6  & 40.4$\pm$1.9  &  46.0$\pm$1.6  &  52.8$\pm$5.0  &  43.8$\pm$4.4  &  \textbf{47.8$\pm$1.1}    \\     
    val\_usefulness       & 59.8$\pm$2.5  & 59.2$\pm$2.7  &  59.4$\pm$1.7  &  61.2$\pm$2.2  &  61.2$\pm$1.9  &  \textbf{61.2$\pm$0.8}    \\     
    val\_knowledge        & 59.6$\pm$0.9  & 44.4$\pm$4.3  &  50.8$\pm$2.6  &  60.4$\pm$4.0  &  44.4$\pm$3.8  &  \textbf{51.0$\pm$2.2}    \\     
    emo\_disgust          & 60.6$\pm$2.9  & 62.8$\pm$2.0  &  61.6$\pm$1.7  &  63.2$\pm$2.2  &  61.2$\pm$2.6  &  \textbf{62.2$\pm$1.3}    \\     
    emo\_trust            & 50.2$\pm$2.8  & 50.2$\pm$3.0  &  50.2$\pm$2.3  &  53.2$\pm$3.8  &  50.8$\pm$2.9  &  \textbf{52.0$\pm$1.6}   \\     
    emo\_surprise         & 45.2$\pm$8.1  & 22.2$\pm$4.4  &  \textbf{29.4$\pm$4.8}  &  40.2$\pm$8.9  &  21.8$\pm$4.1  &  28.4$\pm$5.7    \\     
    emo\_anger            & 68.8$\pm$0.8  & 72.2$\pm$2.3  &  70.6$\pm$0.9  &  69.0$\pm$0.7  &  73.2$\pm$1.6  &  \textbf{71.2$\pm$0.8}    \\     
    pol\_strong\_positive & 59.2$\pm$3.3  & 52.8$\pm$2.8  &  55.8$\pm$2.0  &  61.0$\pm$3.8  &  52.0$\pm$3.5  &  \textbf{56.0$\pm$2.5}    \\     
    pol\_weak\_positive   & 54.8$\pm$1.3  & 55.4$\pm$3.4  &  55.2$\pm$1.8  &  57.4$\pm$2.5  &  56.8$\pm$2.5  &  \textbf{57.4$\pm$0.9}    \\     
    pol\_strong\_negative & 56.0$\pm$4.1  & 64.2$\pm$2.0  &  60.0$\pm$2.0  &  57.2$\pm$1.5  &  64.2$\pm$2.8  &  \textbf{60.4$\pm$1.1}    \\     
    pol\_weak\_negative   & 54.8$\pm$0.8  & 65.0$\pm$2.5  &  59.2$\pm$0.8  &  56.0$\pm$1.9  &  67.0$\pm$2.2  &  \textbf{61.0$\pm$0.7}    \\     
    pol\_ambivalent       & 53.0$\pm$1.2  & 36.2$\pm$2.5  &  43.0$\pm$1.2  &  50.8$\pm$3.8  &  38.0$\pm$2.6  &  \textbf{43.6$\pm$1.5}    \\     
    pol\_neutral          & 88.2$\pm$0.4  & 95.2$\pm$0.8  &  91.6$\pm$0.5  &  89.4$\pm$0.5  &  94.4$\pm$0.9  &  \textbf{91.8$\pm$0.4}    \\     
\hline    
    micro                 & 67.4$\pm$1.1  & 68.4$\pm$0.9  &  67.8$\pm$0.8  &  67.6$\pm$1.3  &  69.2$\pm$1.5  &  \textbf{68.4$\pm$0.5}    \\     
    macro                 & 58.6$\pm$1.3  & 54.2$\pm$1.1  &  55.6$\pm$0.5  &  58.8$\pm$1.3  &  55.2$\pm$1.5  &  \textbf{56.4$\pm$0.5}    \\     
    weighted              & 66.8$\pm$1.1  & 68.4$\pm$0.9  &  67.0$\pm$0.7  &  67.4$\pm$1.1  &  69.2$\pm$1.5  &  \textbf{68.2$\pm$0.4}    \\     
\bottomrule
\end{tabular}
\end{adjustbox}

\end{table}

\subsection{Non-propagated vs Propagated Configurations}

Table~\ref{tab:results3} compares non-propagated (HSSE/MSSE) and propagated (HD/MD) configurations. Utilizing the deep classifier (CLDNSP) leads to a significant performance improvement, regardless of the type of embeddings used (monolingual or cross-lingual).

\begin{table}

\centering
\caption{
     Comparison of HSSE vs. HD and MSSE vs. MD configurations. \textbf{Bold} values indicate the best performance among all architectures.\label{tab:results3}
}
\begin{adjustbox}{width=\columnwidth}
\begin{tabular}{lcccccc}
    \toprule
    \multicolumn{1}{c}{Avg.type}                    & \multicolumn{3}{c}{HSSE} & \multicolumn{3}{c}{HD}  \\
       & P           & R           & F1           & P & R & F1 \\
    \midrule   
    micro          &     66.8$\pm$1.1 & 67.8$\pm$0.8 & 67.4$\pm$0.5 & \textbf{67.4$\pm$1.1} & \textbf{68.4$\pm$0.9} & \textbf{67.8$\pm$0.8}    \\     
    macro           &    57.4$\pm$1.1 & 53.4$\pm$1.1 & \textbf{54.8$\pm$0.8} & \textbf{58.6$\pm$1.3} & 54.2$\pm$1.1 & \textbf{55.6$\pm$0.5}    \\     
    weighted         &   66.2$\pm$0.8 & 67.8$\pm$0.8 & 66.8$\pm$0.8 & \textbf{66.8$\pm$1.1} & \textbf{68.4$\pm$0.9} & \textbf{67.0$\pm$0.7}    \\     
        \midrule   
                        & \multicolumn{3}{c}{MSSE} & \multicolumn{3}{c}{MD}  \\  
           & P           & R           & F1           & P & R & F1 \\
            \midrule   
    micro          &     67.6$\pm$1.9 & 68.2$\pm$1.3 & 68.0$\pm$0.7 & 67.6$\pm$1.3 & \textbf{69.2$\pm$1.5} & \textbf{68.4$\pm$0.5}    \\     
    macro          &     \textbf{59.0$\pm$1.9} & 54.4$\pm$1.5 & 55.6$\pm$0.5 & 58.8$\pm$1.3 & \textbf{55.2$\pm$1.5} & \textbf{56.4$\pm$0.5}    \\     
    weighted       &     67.4$\pm$1.7 & 68.2$\pm$1.3 & 67.6$\pm$0.5 & 67.4$\pm$1.1 & \textbf{69.2$\pm$1.5} & \textbf{68.2$\pm$0.4}    \\             
\bottomrule
\end{tabular}
\end{adjustbox}

\end{table}    

\section{Conclusions}
\label{sec:conclusions}

 Our study highlights the effectiveness of MSSE and CLDNSP, our proposed methods for sentiment and emotion propagation in multilingual WordNets. We found that the MD configuration, which uses both MSSE and CLDNSP, outperforms other configurations in terms of performance metrics. This emphasizes the significance of integrating cross-lingual information and deep neural networks in sentiment propagation tasks. Furthermore, our results indicate that our propagation method is highly effective, as it delivers considerable gains regardless of the embeddings used.

As part of further research, we recommend two steps. Firstly, we suggest extending the MSSE embedding in a similar manner as the HSSE method, but this time using multilingual aligned embeddings \cite{joulin2018loss}. Secondly, we recommend testing the impact of the propagated dataset on the quality of sentiment recognition in text.

\section*{Acknowledgements}
This work was financed by 
(1) contribution to the European Research Infrastructure "CLARIN ERIC - European Research Infrastructure Consortium: Common Language Resources and Technology Infrastructure", 2022-23 (CLARIN Q);
(2) the European Regional Development Fund, as a part of the 2014-2020 Smart Growth Operational Programme, projects no. POIR.04.02.00-00C002/19, POIR.01.01.01-00-0923/20, POIR.01.01.01-00-0615/21, and POIR.01.01.01-00-0288/22; 
(3) the statutory funds of the Department of Artificial Intelligence, Wroclaw University of Science and Technology;
(4) the Polish Ministry of Education and Science within the programme “International Projects Co-Funded”;
(5) the European Union under the Horizon Europe, grant no. 101086321 (OMINO). However, the views and opinions expressed are those of the author(s) only and do not necessarily reflect those of the European Union or the European Research Executive Agency. Neither the European Union nor European Research Executive Agency can be held responsible for them.

\bibliography{main}

\end{document}